\documentclass[]{article}
\pdfoutput=1
\usepackage[space]{grffile}
\usepackage{caption}
\usepackage{subcaption}
\usepackage{multirow}
\usepackage{url}
\usepackage{algorithmic}
\usepackage{amsmath,amsfonts,amssymb}
\usepackage{graphicx}
\usepackage[dvipsnames]{xcolor}
\usepackage{cite}
\usepackage{geometry}
\usepackage{subcaption}
\usepackage{textcomp}
\usepackage{makecell}
\usepackage{hyperref}
\usepackage{authblk}
\usepackage[maxfloats=200]{morefloats}
\title{Problems with automating translation of movie/TV show subtitles}
\author{Prabhakar Gupta}
\author{Mayank Sharma}
\author{Kartik Pitale}
\author{Keshav Kumar}
\affil{Amazon}
\affil[ ]{\textit {\{prabhgup,mysharm,kppitale,kumakesh\}@amazon.com}}

\begin{document}
\date{}
\maketitle

\begin{abstract}
We present 27 problems encountered in automating the translation of movie/TV show subtitles. We categorize each problem in one of the three categories \textit{viz.} problems directly related to textual translation, problems related to subtitle creation guidelines, and problems due to adaptability of machine translation (MT) engines. We also present the findings of a translation quality evaluation experiment where we share the frequency of 16 key problems. We show that the systems working at the frontiers of Natural Language Processing do not perform well for subtitles and require some post-processing solutions for redressal of these problems.
\end{abstract}

\section{Introduction}
Subtitling a video enhances the audio-visual experience. It helps viewers watch content in languages in which they lack proficiency. With over 450 million hearing impaired people across the globe\footnote{World Health Organization: \url{https://www.who.int/news-room/fact-sheets/detail/deafness-and-hearing-loss}}, subtitling broadens the reach of companies in the multimedia domain like Prime Video. Subtitles also aid in better understanding of inaudible spoken words (like whispering), a person talking in a different accent/language, background noises etc. Thus, a correct subtitle is imperative for better viewing experience.  
Even now, Prime Video mostly does manual subtitle translation. This process is time consuming (average of 20 hours per hour of content), expensive (average \$12 per minute of content), not scalable due to constant catalog growth across languages, relies on subjective knowledge of translators and for organizations dealing with sensitive data, it also limits usage. One possible solution is to automate the process. Statistical Machine Translation (SMT) systems \cite{Brown1990ASA} have been around for years but they have not been able to outperform humans in generating a natural-sounding translation. Authors in \cite{Volk2009TheAT, Sennrich2010MachineTO, Mller2013StatisticalMT, prabhakar2019Unsupervised} have tried to identify some problems with automated subtitle translation, however, their work was restricted to SMT. With progress in the domain of Natural Language Processing (NLP) and advent of Deep Learning (Neural Machine Translation (NMT)  \cite{Hieber2017SockeyeAT,45610,Luong2015EffectiveAT, Britz2017MassiveEO,Vaswani2017AttentionIA}) in recent years, we have now started exploring solutions for the automated translation of subtitles.

In this work, we list (\textbf{not exhaustive}) and explain the problems we discovered during our research\footnote{There are resources/frameworks like Multidimensional Quality Metrics (MQM) framework which provide metrics for translation quality estimation but they are generally used by human evaluators as a ``checklist" to ensure translation quality. We could not find any resources discussing the automated process for the same.} while generating automated translations. We classify each problem into one of three categories; firstly, the problems directly related to textual translation, secondly, problems related to subtitle creation guidelines, and lastly, problems due to adaptability of MT engines. Some of these problems can be solved using post-processing of the MT output like the incorrect spacing errors, incorrect spellings, addition/deletion of words while others like language and cultural nuances require sophisticated solutions which include building better MT engines and proper understanding of language. While listing the possible solutions to individual problems is not the focus of the paper, it gives an insight into the type of solutions which can be devised to tackle each problem.

This paper is divided into three major sections. Section \ref{sec:Problems} elucidates the key problems in subtitle translation using MT systems. As a running example, we majorly focus on the problems in English to German subtitle translation with some exceptions. Section \ref{sec: Pilotprogram} describes the Subtitle Validation Experiment we conducted to validate the key identified problems, and present a corrected solution to the output of the MT engine. The idea is to identify the gravitas of individual problems in terms of understanding of translated text, readability and the frequency of occurrence. Finally, with section \ref{sec:Conclusion} we conclude the paper, with the disclaimer that though the list provided in current paper is not exhaustive, it describes the key problems in automated subtitle translation during our experiments. Problems that arise with dubbed content, generating captions, audio transcription and corrupt source text are outside the scope of this work.

\section{Problems}\label{sec:Problems}
\begin{figure}[hbtp]
    \centering
    \includegraphics[width=0.3\textwidth]{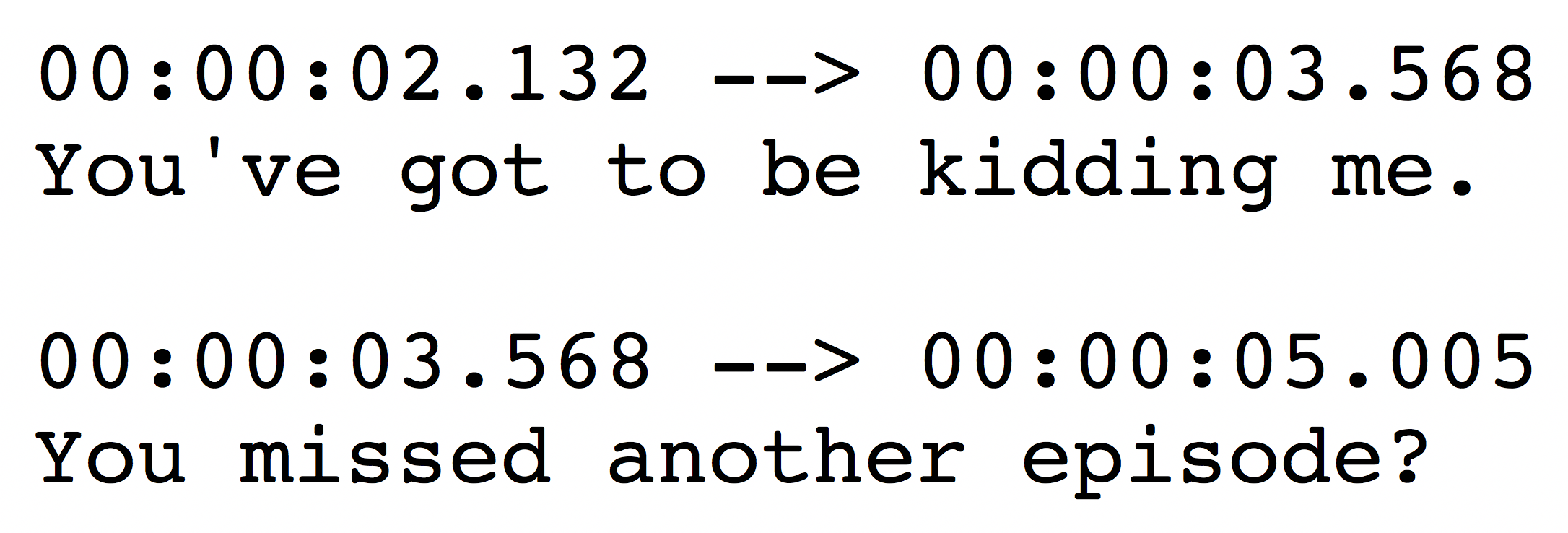}
    \caption{Subtitle Example}
    \label{fig:subtitle_example}
\end{figure}

A typical subtitle block consists of two timestamps and one text block. Figure \ref{fig:subtitle_example} presents an example of a subtitle file with two subtitle blocks in the VTT format\footnote{\url{https://en.wikipedia.org/wiki/WebVTT}}. Timestamps define the period in which the text block is to be shown and the timestamp structure depends on the file format. Different content publishers\footnote{BBC Subtitle Guidelines: \url{http://bbc.github.io/subtitle-guidelines}} have different guidelines for onscreen subtitle creation process to maintain a uniform viewing experience. While automating, MT systems can violate these guidelines.
We now enumerate the key problems encountered in the automated translation of subtitles




\subsection{Problems related to subtitle creation guidelines}
\begin{enumerate}
    \item \textbf{Repeated phrases/words}: \label{repeatedphraseswords}
    MT engines literally translate the repeated words or phrases in the source sentence  \cite{le2017improving}. As shown in Table \ref{table:repeatedphraseswords}, the repetitions are sometimes removed by humans only keeping the first occurrence to avoid unnecessary increase in length of translation.
    \begin{table}[hbtp]
    	\begin{tabular}{|p{1cm}|p{5cm}|p{5cm}|p{5cm}|}
    		\hline
    		\textbf{\#} & \textbf{Source Subtitle} & \textbf{Machine Translated Subtitle}  & \textbf{Human Translated Subtitle}  \\ \hline
    		1 & so it's very, very frustrating. & also ist es sehr, \textcolor{red}{sehr} frustrierend. & also ist es äußerst frustrierend. \\ \hline
    		2 & Go, go, go! & Los, \textcolor{red}{los, los}! & Los! \\ \hline
    		3 & Get over there! Hurry up! Get over there. & Geh da rüber! Beeil dich! \textcolor{red}{Geh da rüber}. & Geht da rüber! Beeilt euch! \\ \hline
    		4 & Yes, yes. & Ja, \textcolor{red}{ja}. & Ja. \\ \hline
    	\end{tabular}
    	\caption{Repeated phrases/words}
    	\label{table:repeatedphraseswords}
    \end{table}
    
    \item \textbf{Compound words}:
    Compound words can either be in closed (like \textit{firefly}, \textit{softball}), open (like \textit{ice cream}) or hyphenated (like \textit{father-in-law}). Understanding and translating small and frequent compound words is easy. However, German is notorious for lengthy compound words   \cite{Marco2017SimpleCS}, and a lot of research has gone into solving this problem for SMT  \cite{Popovic2006StatisticalMT, Escartn2014GermanCA}. These long compound words are mostly absent from the vocabulary of MT systems resulting in poor translation.
    
    \item \textbf{Incorrect spacing error}:
    If an incorrect spacing in present around punctuation marks, hyphens or places where space or lack thereof changes the interpretation, it is classified as an incorrect spacing error. As shown in Table \ref{table:incorrectspacingerror}, it is necessary to follow correct spacing after hyphens and ellipses. 
    \begin{table}[hbtp]
    	\begin{tabular}{|p{1cm}|p{5cm}|p{5cm}|p{5cm}|}
    		\hline
    		\textbf{\#} & \textbf{Source Subtitle} & \textbf{Machine Translated Subtitle}  & \textbf{Human Translated Subtitle}  \\ \hline
    		1 & -Thank you. -Oh, boy. & - Danke. - Oh, Junge. & -Danke. -Oh, Junge. \\ \hline
    		2 & ...and a silver Sebright hen. & ... und eine silberne Sebright-Henne. & ...und eine silberne Sebright-Henne. \\ \hline
    		3 & ...that there was more water in the system. & ... dass es mehr Wasser im System gab. & ...dass es mehr Wasser im System gab. \\ \hline
    	\end{tabular}
    	\caption{Incorrect Spacing Error}
    	\label{table:incorrectspacingerror}
    \end{table}
    

    \item \textbf{Inconsistent translation of non-text characters}:
    Symbols like hyphens, line breaks (\textbackslash n), HTML tags, etc. are introduced in subtitles to provide additional information or dictate how text is rendered on-screen. These symbols have to be removed before translating the subtitle text and are to be added back in translation output. However, it is difficult to identify the correct positions where these symbols need to be inserted. As shown in Table \ref{table:Inconsistenttranslationofnontextcharacters}, HTML tags (mainly italics) are used to signify that the speaker is off-screen. Hyphens with line breaks are used to indicate presence of multiple speakers. 
    \begin{table}[hbtp]
    	\begin{tabular}{|p{1cm}|p{5cm}|p{5cm}|p{5cm}|}
    		\hline
    		\textbf{\#} & \textbf{Source Subtitle} & \textbf{Machine Translated Subtitle}  & \textbf{Human Translated Subtitle}  \\ \hline
    		1     & \textless i\textgreater that lurked beneath everyday palace life.\textless /i\textgreater & die unter dem alltäglichen Palastleben lauerten. & \textcolor{red}{\textless i\textgreater} das im alltäglichen Palastleben lauerte.\textcolor{red}{\textless /i\textgreater}\\ \hline
    		2     & But it could also be short for \textless i\textgreater specularius,\textless /i\textgreater & Aber es könnte auch kurz Forspekularius sein, & Es könnte auch kurz für \textcolor{red}{\textless i\textgreater} specularius\textcolor{red}{\textless /i\textgreater}  sein,\\ \hline
    		3     & that swirl around the undersea ledges \textbackslash n and mountains. & , die sich um die Unterwasservorsprünge und Berge drehen. & die um die Vorsprünge und Berge \textcolor{red}{\textbackslash n} unter Wasser wirbeln. \\ \hline
    		
    	\end{tabular}
    	\caption{Inconsistent translation of non-text characters}
    	\label{table:Inconsistenttranslationofnontextcharacters}
    \end{table}

    
    \item \textbf{Mixed languages in a movie}:
    Some movies have more than one primary language. It is critical to identify the different language blocks and pass them through the right translation models for the correct output \cite{lakew2018neural}. For example, Babel\footnote{\url{www.imdb.com/title/tt0449467}} has characters speaking in English, Arabic, Spanish and Japanese. A Bollywood movie, Chennai Express\footnote{\url{www.imdb.com/title/tt2112124}} has both Hindi and Tamil speaking characters.
    
    \item \textbf{Subtitle block count integrity}:
    During translation, an MT engine translates each subtitle block individually. A human translator on the other hand, might change the number of subtitle blocks. This primarily happens due to one of three reasons; firstly, either the number of words changed significantly and the translator felt the need to merge adjacent subtitle blocks or split one block into two or more. Secondly, a subtitle block on it's own did not make much sense and translator considered more than one block for translation. Finally, subtitle time should be under a certain reading speed (number of words per second). As shown in Table \ref{table:subtitleblockcountintegrity}, a human translator would split/merge a subtitle block, but it will be difficult for the translation engine to determine the exact point where a block is to be split, or identify which blocks need to be merged. 
    \begin{table}[hbtp]
    	\begin{tabular}{|p{1cm}|p{5cm}|p{5cm}|p{5cm}|}
    		\hline
    		\textbf{\#} & \textbf{Source Subtitle} & \textbf{Machine Translated Subtitle}  & \textbf{Human Translated Subtitle}  \\ \hline
    		1     & One cry meant you were hungry... & Ein Schrei bedeutete, dass du Hunger hattest... & Ein Schrei bedeutet, ich habe Hunger. \\ \hline
    		2     & -It's a boy. & - Es ist ein Junge. & -Es ist ein Junge! -Wie geht es ihm? Ist er okay? \\ \hline
    		3     & How is he? Is he okay? & Wie geht es ihm? & \\ \hline
    	\end{tabular}
    	\caption{Subtitle block count integrity}
    	\label{table:subtitleblockcountintegrity}
    \end{table}

\end{enumerate}

\subsection{Problems related to textual translation}
\begin{enumerate}
    \item \textbf{Paraphrased translations}:
    For each language, there is a specific word length and character length rule to offer the best viewing experience. In some cases, a less than ideal translation will have to be used to meet these strict rules. For example, for German, Prime Video uses 42 characters per line, 3 lines per block. 
    In other case, the differences in speech conventions between source and target language can cause humans to paraphrase the sentences. As shown in Table \ref{table:paraphrasing}, an automated translation engine tries to give out a literal translation that disregards the guidelines whereas a human can provide an imperfect/alternate translation to adhere to these rules \cite{romero2009more}.
    \begin{table}[hbtp]
        \centering
        \scalebox{0.75}{
    	\begin{tabular}{|p{1cm}|p{5cm}|p{1.2cm}|p{5cm}|p{1.2cm}|p{5cm}|p{1.2cm}|}
    		\hline
    		\textbf{\#} & \textbf{Source Subtitle} & \textbf{Length} & \textbf{Machine Translated Subtitle} & \textbf{Length} & \textbf{Human Translated Subtitle} &  \textbf{Length} \\ \hline
    		1 & Come by and drive it whenever you want. & 39 & Kommen Sie vorbei und fahren Sie es, wann immer Sie wollen. & 59 & Komm jederzeit zum Fahren vorbei. & 33  \\ \hline
    		2 & I chose to hide it from everyone. & 33 & Ich habe mich entschieden, es vor allen zu verstecken. & 54 & entschied ich mich, es zu verstecken. & 37 \\ \hline
    		3 & with the men guilty of those crimes. & 36 & mit den Männern, die sich dieser Verbrechen schuldig gemacht haben. & 67 & mit den Schuldigen dieser Verbrechen. & 37 \\ \hline
    		4 & Nothing but an ordinary match folder. & 37 & Nichts als ein gewöhnlicher Übereinstimmungsordner. & 51 & Nur ein gewöhnlicher Streichholzbrief. & 38 \\ \hline
    	\end{tabular}
    	}
    	\caption{Paraphrased Translations}
    	\label{table:paraphrasing}
    \end{table}
    
    
    
    \item \textbf{Translating Idioms}:
    Literal translation of an idiom does not make much sense. Identifying idioms is very difficult on its own and their translation is even more challenging  \cite{Anastasiou2010IdiomTE}. There may not be an equivalent idiom/phrase in the target language. Even if it exists, correctly fitting it within the context with grammatical accuracy is difficult. For example, the German idiom, \textit{``Da kannst du Gift drauf nehmen"} literally means \textit{``you can take poison on that"} but it's equivalent English idiom is \textit{``You can bet your life on that"}

    \item \textbf{Literal Translation v/s Contextual Translation}:
    A single phrase can convey various meanings depending on the framing context as shown in Table  \ref{table:literaltranslationvsretentionofmeaning}. Automated translation engines don’t do well for such cases \cite{knowles2018context}. For example, the phrase ``beats me" can have a different meaning based on the context. 
    
    \begin{table}[hbtp]
    	\begin{tabular}{|p{1cm}|p{5cm}|p{5cm}|p{5cm}|}
    		\hline
    		\textbf{\#} & \textbf{Source Subtitle} & \textbf{Machine Translated Subtitle}  & \textbf{Human Translated Subtitle}  \\ \hline
    		1 & But it only lasted four hours. & Aber \textcolor{red}{es dauerte} nur vier Stunden. & Aber sie hielt nur vier Stunden. \\ \hline
    		2 & He said, ``Get us out of here. We\textquotesingle re stinking." & Er sagte: ``Bringen Sie uns hier raus. Wir \textcolor{red}{stinken}." & Er sagte: ``Bringen Sie uns hier raus. Wir ertrinken." \\ \hline
    	\end{tabular}
    	\caption{Literal Translation v/s Contextual Translation}
    	\label{table:literaltranslationvsretentionofmeaning}
    \end{table}
    
    \item \textbf{Profanity}:
    Movies, being an artistic medium, may have cuss words or derogatory phrases that have adapted versions in different languages. During translation, the same level of profanity should be maintained. It is not always possible to find a correct translation of some profane words and phrases. For example, a language might consider a phrase/word as derogatory while it's literal translation in some other language might be acceptable. The insult from Pulp Fiction\footnote{\url{www.imdb.com/title/tt0110912}}, \textit{``fucking asshole"} is translated by an MT engine to \textit{``puto gilipollas"} in Spanish which means \textit{``asshole"} whereas it was translated to \textit{``cabrón"} by human translators meaning \textit{``dumbass"} conveying a rather similar meaning \cite{vilaCabrera2015AnAO}.
    
    
    \item \textbf{Identify text not to translate}:
    In certain cases, parts of a sentence should be excluded from translation/transliteration\footnote{Transliteration is the process of transferring a word from the alphabet of one language to another}. 
    As shown in Table \ref{table:identifytextnottotranslate}, \textit{``MARY BEARD"} was not identified as a proper noun and was translated like a common noun. Human translators use translation memories called Key Names and Phrases (KNPs)\footnote{Dictionary of translations of common phrases and names to have consistent translation across movie/TV series} to identify such text along with their own judgment which lacks in an MT engine.
    \begin{table}[hbtp]
    	\begin{tabular}{|p{1cm}|p{5cm}|p{5cm}|p{5cm}|}
    		\hline
    		\textbf{\#} & \textbf{Source Subtitle} & \textbf{Machine Translated Subtitle}  & \textbf{Human Translated Subtitle}  \\ \hline
    		1 & CALIGULA WITH MARY BEARD & CALIGULA MIT \textcolor{red}{MARY BART} & CALIGULA MIT MARY BEARD \\ \hline
    	\end{tabular}
    	\caption{Identify text not to translate}
    	\label{table:identifytextnottotranslate}
    \end{table}
    
    
    
    \item \textbf{Addition/Omission of words}:
    In a translated subtitle block, words can be added or removed from source during translation. As shown in Table \ref{table:additionomission}, addition or omission of the words can be very trivial but in some cases it can alter the meaning of the sentence or provide incomplete information to the viewer.
    \begin{table}[hbtp]
    	\begin{tabular}{|p{1cm}|p{5cm}|p{5cm}|p{5cm}|}
    		\hline
    		\textbf{\#} & \textbf{Source Subtitle} & \textbf{Machine Translated Subtitle}  & \textbf{Human Translated Subtitle}  \\ \hline
    		1 & world-wide problems that go beyond & weltweite Probleme, die über & \textcolor{red}{um} weltweite Probleme, die über  \\ \hline
    		2 & in the direction of world government for the Antichrist. & in Richtung der Weltregierung für den Antichristen. & in Richtung der Weltregierung für den Antichristen \textcolor{red}{steuern}.  \\ \hline
    		3 & The Trilateral Commission is widely seen & Die Trilaterale Kommission \textcolor{red}{ist} weit verbreitet & Die Trilaterale Kommission wird allgemein  \\ \hline
    	\end{tabular}
    	\caption{Addition/Omission of words}
    	\label{table:additionomission}
    \end{table}
    
    \item \textbf{Word Order Error}:
    In translated subtitle, it is possible that MT engine can introduce error in form of word order. For example, If the intended sequence of words was A-B-C, and the translation comes out to be B-A-C --- this can result in a grammatical error or can alter the meaning of sentence. As shown in Table \ref{table:wordordererrors}, in some cases, a couple of words can get swapped or the position of a could be incorrect.
    \begin{table}[hbtp]
    	\begin{tabular}{|p{1cm}|p{5cm}|p{5cm}|p{5cm}|}
    		\hline
    		\textbf{\#} & \textbf{Source Subtitle} & \textbf{Machine Translated Subtitle}  & \textbf{Human Translated Subtitle}  \\ \hline
    		1 & this is the hottest planet in the solar system. & \textcolor{red}{das ist} der heißeste Planet im Sonnensystem. & \textcolor{red}{ist das} der heißeste Planet im Sonnensystem.  \\ \hline
    		2 & the amount of CO2 in the atmosphere has increased nearly 40\%, & die CO2-Menge in der Atmosphäre \textcolor{red}{hat} fast 40\% zugenommen, & \textcolor{red}{hat} die CO2-Menge in der Atmosphäre
    fast 40\% zugenommen, \\ \hline
    	\end{tabular}
    	\caption{Word Order Error}
    	\label{table:wordordererrors}
    \end{table}
    
    \item \textbf{Language nuances}:
    In German, \textit{``Du"} is used with people that are very well known and \textit{``Sie"} is used with unfamiliar people. In French, the choice of pronoun \textit{``tu"} and \textit{``vous"} is matter of etiquette. \textit{``tu"} is used for singular informal and \textit{``vous"} is plural and/or formal. Choosing the wrong pronoun can have negative consequences. In Japanese, second person pronouns are rarely used --- even if the speaker is in front of the person he/she is referring to, it is more common to address them using their family name. As shown in Table \ref{table:formalinformaltranslationinconsistency}, automated translation uses incorrect pronouns which were corrected by human translators.
    \begin{table}[hbtp]
    	\begin{tabular}{|p{1cm}|p{5cm}|p{5cm}|p{5cm}|}
    		\hline
    		\textbf{\#} & \textbf{Source Subtitle} & \textbf{Machine Translated Subtitle}  & \textbf{Human Translated Subtitle}  \\ \hline
    		1 & Throw cake at the clown. & \textcolor{red}{Werfen Sie} Kuchen auf den Clown. & \textcolor{red}{Werft} Kuchen auf den Clown. \\ \hline
    		2 & Everybody's waiting to congratulate you. & Alle warten darauf, \textcolor{red}{Ihnen} zu gratulieren. & Alle warten darauf, \textcolor{red}{dir} zu gratulieren. \\ \hline
    		3 & ``I would like it to pass on to you & ``Ich möchte, \textcolor{red}{dass es an Sie} weitergibt & ``Ich möchte \textcolor{red}{sie an dich} weitergeben, \\ \hline
    	\end{tabular}
    	\caption{Language nuances}
    	\label{table:formalinformaltranslationinconsistency}
    \end{table}

    \item \textbf{Agreement Error}:
    Agreement error occurs when one or more target words disagree in any form of inflection\footnote{Inflection is a change in the form of a word (generally the ending) to express attribute such as tense, mood, person, number, case, and gender.}. As shown in Table \ref{table:agreementerror}, it can change the meaning of a sentence and create confusion for the viewer.
    \begin{table}[hbtp]
    	\begin{tabular}{|p{1cm}|p{5cm}|p{5cm}|p{5cm}|}
    		\hline
    		\textbf{\#} & \textbf{Source Subtitle} & \textbf{Machine Translated Subtitle}  & \textbf{Human Translated Subtitle}  \\ \hline
    		1 & he was the great-grandson of Augustus, & er war der Urenkel Augustus, & er war der Urenkel des Augustus, \\ \hline
    		2 & who held the reins of power. & der die Zügel der Macht hielt. & die Zügel der Macht in den Händen hielt. \\ \hline
    		3 & But how much higher is it? & Aber wie viel höher ist es? & Aber wie viel höher ist sie? \\ \hline
    	\end{tabular}
    	\caption{Agreement Error}
    	\label{table:agreementerror}
    \end{table}
    
    \item \textbf{Misspelling}:
    The effect of misspelling on MT quality is widely known  \cite{Galinskaya2014MeasuringTI}. In subtitles, we mark a misspelling when it violates the movie-specific glossary or that of the target language. As shown in Table \ref{table:misspelling}, misspelling can alter the meaning of translation.
    \begin{table}[hbtp]
    	\begin{tabular}{|p{1cm}|p{5cm}|p{5cm}|p{5cm}|}
    		\hline
    		\textbf{\#} & \textbf{Source Subtitle} & \textbf{Machine Translated Subtitle}  & \textbf{Human Translated Subtitle}  \\ \hline
    		1 & on small pieces of limestone, on ostraca. & auf kleinen Stücken Kalkstein, auf \textcolor{red}{Ostraca}. & auf kleinen Stücken Kalkstein, auf Ostraka. \\ \hline
    		2 & The look in his eyes-- & Der Blick in \textcolor{red}{seine} Augen... & Der Blick in seinen Augen... \\ \hline
    		3 & all at once. & \textcolor{red}{alle} auf einmal. & alles auf einmal. \\ \hline
    		4 & that accident. & \textcolor{red}{diesen} Unfall. & dieser Unfall. \\ \hline
    	\end{tabular}
    	\caption{Misspelling}
    	\label{table:misspelling}
    \end{table}
    
    \item \textbf{Nonsensical Translation Error}:
    Errors that occur due to an incorrect translation or incomprehensible translation. As shown in Table \ref{table:nonsensicaltranslationerror}, human translation is very different from automated translation since MT engine either literally translated the sentence or did not translate the sentence correctly.
    \begin{table}[hbtp]
    	\begin{tabular}{|p{1cm}|p{5cm}|p{5cm}|p{5cm}|}
    		\hline
    		\textbf{\#} & \textbf{Source Subtitle} & \textbf{Machine Translated Subtitle}  & \textbf{Human Translated Subtitle}  \\ \hline
    		1 & you've got to walk in their footsteps. & Sie müssen in ihre Fußstapfen gehen. & muss man in ihre Fußstapfen treten. \\ \hline
    		2 & we have an instant gateway, & wir haben ein sofortiges Tor, & schaffen wir uns einen Zugang \\ \hline
    		3 & and his hip flask. Everything is there. & und seinen Hüftkolben. Alles ist da. & und seine Hüftflasche. Alles ist da. \\ \hline
    	\end{tabular}
    	\caption{Nonsensical Translation Error}
    	\label{table:nonsensicaltranslationerror}
    \end{table}
    
    \item \textbf{Not-translated words}:
    During automated translation, for an out-of-vocabulary (OOV) word, the MT engine might consider it as a proper noun and choose not translate it  \cite{Luong2015AddressingTR}. As shown in Table \ref{table:nottranslatedwords}, there are some words that were not translated but should have been translated.
    \begin{table}[hbtp]
    	\begin{tabular}{|p{1cm}|p{5cm}|p{5cm}|p{5cm}|}
    		\hline
    		\textbf{\#} & \textbf{Source Subtitle} & \textbf{Machine Translated Subtitle}  & \textbf{Human Translated Subtitle}  \\ \hline
    		1 & The true school for Che's New Man & Die wahre Schule für Ches \textcolor{red}{New Man} & Die wahre Schule für Ches Neuen Menschen \\ \hline
    		2 & It was a kind of paean, & Es war eine Art \textcolor{red}{Paean}, & Es war eine Art Lobgesang, \\ \hline
    		3 & They all have their little quirks. & Sie haben alle ihre kleinen \textcolor{red}{Quirks}. & Sie haben alle ihre kleinen Eigenarten. \\ \hline
    	\end{tabular}
    	\caption{Not-translated words}
    	\label{table:nottranslatedwords}
    \end{table}
    
    \item \textbf{Over-Translation Error}:
    Errors due to translation being more specific than required \cite{Tu2016ModelingCF}. For example, source text talks about a woman and MT engine uses a term suitable for an older woman instead of more generic one.
    
    \item \textbf{Translating stammering}:
    Stammering occurs because a character might be nervous or might have a speech defect. For an English sentence, \textit{``I w\ldots w\ldots was going there"}, typical MT system outputs \textit{``Ich wollte da hingehen."} which means \textit{``I wanted to go there."} which is an incorrect translation. Identifying a case of stammering and translating it is a difficult problem \cite{scripture1922some} because it depends on how it should be translated. The above text can be translated as \textit{``[stammers] Ich ging dort hin"} or \textit{``Ich g\dots g\ldots ging dort hin"} or by completely ignoring the stammering part \textit{``Ich ging dort hin"}. 

\end{enumerate}
\subsection{Machine Translation adaptability Problems}
\begin{enumerate}
    \item \textbf{Cultural nuances}:
    A language spoken across countries (or locales) can contain different words to represent a concept. For example, a \textit{cookie} in US is called a \textit{biscuit} in the UK, \textit{petrol/fuel} in other countries translates to \textit{gas} in the US. Other cases of such languages include Castilian Spanish v/s Mexican Spanish, Portuguese in Portugal v/s Brazil, and  Hinglish\footnote{Hinglish is a language that combines words from English and South Asian languages like Hindi}. For example, if in an Indian movie someone says \textit{``jump off Qutub Minar"}, it is easy to understand for Indian audiences to relate that \textit{Qutub Minar} is long tower-like structure but if we want to translate this for French audience, they would be able to relate better to an \textit{Eiffel Tower} reference.

    
    \item \textbf{Wrong Lexical Translation}:
    Errors that occur because word/phrase like abbreviation and acronym is incorrectly translated. As shown in Table \ref{table:wronglexicaltranslation}, the error can occur if a translated lexicon violates glossary, standard language, industry usage, is inconsistent with other translations of the source term or denotes a concept different from the source term.
    \begin{table}[hbtp]
    	\begin{tabular}{|p{1cm}|p{5cm}|p{5cm}|p{5cm}|}
    		\hline
    		\textbf{\#} & \textbf{Source Subtitle} & \textbf{Machine Translated Subtitle}  & \textbf{Human Translated Subtitle}  \\ \hline
    		1 & come from this village, except one. & kommen aus diesem Dorf, außer einem. & kommt aus diesem Dorf, außer einem. \\ \hline
    		2 & So, we go into the front room here, & Also gehen wir hier in den Vorderraum, & Hier gehen wir ins Vorderzimmer, \\ \hline
    		3 & because this area directly adjoins & weil sich dieser Bereich direkt anschließt & denn dieser Bereich schließt direkt \\ \hline
    	\end{tabular}
    	\caption{Wrong Lexical Translation}
    	\label{table:wronglexicaltranslation}
    \end{table}
    
    
    \item \textbf{Grammatical accuracy prioritization}:
    Some movies introduce grammatical inaccuracies to give characters distinguishing traits. For example, in \textit{The Empire Strikes Back}\footnote{\url{www.imdb.com/title/tt0080684}}, when \textit{Yoda}\footnote{\url{www.starwars.com/databank/yoda}} meets \textit{Luke Skywalker}\footnote{\url{www.starwars.com/databank/luke-skywalker}} for the first time, he says - \textit{``Looking? Found someone, you have, I would say, hmmm?"}. This sentence is intended to be grammatically incorrect but will cause problems during translation. It is required for the translations to retain the artistic intent and convey accurate meaning that might not be grammatically correct.
    
    \item \textbf{Word Structure Error}:
    A word structure error occurs when the translation is grammatically and technically correct but uses incorrect morphological form such as case, gender, number, tense, prefix, suffix, infix, etc. As shown in Table \ref{table:wordstructureerror}, the translation text is correct grammatically and follows all technical subtitle specifications but is morphologically incorrect translation.
    \begin{table}[hbtp]
    	\begin{tabular}{|p{1cm}|p{5cm}|p{5cm}|p{5cm}|}
    		\hline
    		\textbf{\#} & \textbf{Source Subtitle} & \textbf{Machine Translated Subtitle}  & \textbf{Human Translated Subtitle}  \\ \hline
    		1 & Although deathly silent today, & Obwohl heute \textcolor{red}{Todesstille} herrscht, & Obwohl heute Totenstille herrscht, \\ \hline
    		2 & gold anklets, but the most exciting thing & \textcolor{red}{Gold Knöchel, aber die aufregendste Sache} & goldene Fußkettchen, aber das Spannendste \\ \hline
    		3 & This is the Great Devourer. & Das ist der große \textcolor{red}{Verschlinger}. & Das ist die große Fresserin Ammit. \\ \hline
    	\end{tabular}
    	\caption{Word Structure Error}
    	\label{table:wordstructureerror}
    \end{table}
    
    \item \textbf{Format Errors}:
    Errors that occur because the numbers or numerals are incorrectly translated. As shown in Table \ref{table:numbersanddatecheck}, the MT output retains the Imperial system when it should have used the International System of Units which is more prevalent in Germany.
    \begin{table}[hbtp]
    	\begin{tabular}{|p{1cm}|p{5cm}|p{5cm}|p{5cm}|}
    		\hline
    		\textbf{\#} & \textbf{Source Subtitle} & \textbf{Machine Translated Subtitle}  & \textbf{Human Translated Subtitle}  \\ \hline
    		1     & which from \textcolor{red}{15,000 feet} must've looked to my bomb aimer like a dinky toy, & die von \textcolor{red}{15.000 Fuß} muss auf meine Bombenauslöser wie ein dinky Spielzeug, & das aus \textcolor{red}{4500 m} Höhe für den Schützen 
    wohl wie ein schäbiges Spielzeug aussah,  \\ \hline
    		2     & even though it was \textcolor{red}{900 feet} long. & obwohl es \textcolor{red}{900 Fuß} lang war. & obwohl es fast \textcolor{red}{300 m} lang war.  \\ \hline
    		3     & and a Mosquito tank of \textcolor{red}{50 gallons}, one on top of the other, & und einen Moskitonistank von \textcolor{red}{50 Gallonen}, einer über dem anderen, & und ein Mosquitotank für fast \textcolor{red}{200 Liter}, einer über dem anderen, \\ \hline
    	\end{tabular}
    	\caption{Format Errors (Metric system errors, date errors, etc.)}
    	\label{table:numbersanddatecheck}
    \end{table}
    
    \item \textbf{Impact of movie genre}:
    Subtitles contain contextual information, hence, different genres adapt differently to MT \cite{Wees2018EvaluationOM}. For example, sarcastic/cultural comedy like The Grand Tour\footnote{\url{www.imdb.com/title/tt5712554}} do not translate as good as a documentary like Aerial America\footnote{\url{www.imdb.com/title/tt2735544}} that mostly contains factual information. The overall genre has an impact on translation quality but at subtitle block/scene level genre affects translation quality even more. For example, a comedy movie may have a mix of scenes/blocks with various genres.
    
    \item \textbf{Invented Languages}:
    Certain movies invent languages for imparting authenticity to character groups. For example, Elvish in the Lord Of The Rings Series\footnote{\url{www.imdb.com/list/ls005053232}} and Dothraki in Game of Thrones\footnote{\url{www.imdb.com/title/tt0944947}}. The subtitles for parts in which an artificial language is spoken may contain text like
    \textit{[``Yer jalan atthirari anni"]} which MT engine cannot translate but a human will translate to \textit{[``Speaking in Dothraki"]}.
        
        
\end{enumerate}

\section{Subtitle Validation Experiment}
\label{sec: Pilotprogram}
In this section, we describe the subtitle validation experiment conducted to identify the frequency of 16 key problems in the automated translation of subtitles from English to six target languages \textit{viz.} German, Chinese (simplified), French, Castilian Spanish, Arabic and Brazilian Portuguese. The experiment was performed on 56 movie subtitle files containing a total of 17,977 subtitle blocks. The English subtitles were generated by humans and target subtitles were generated using an MT system trained using \cite{Hieber2017SockeyeAT}.

\begin{figure*}[t]
    \centering
    \begin{subfigure}[b]{0.30\textwidth}
        \includegraphics[width=\textwidth]{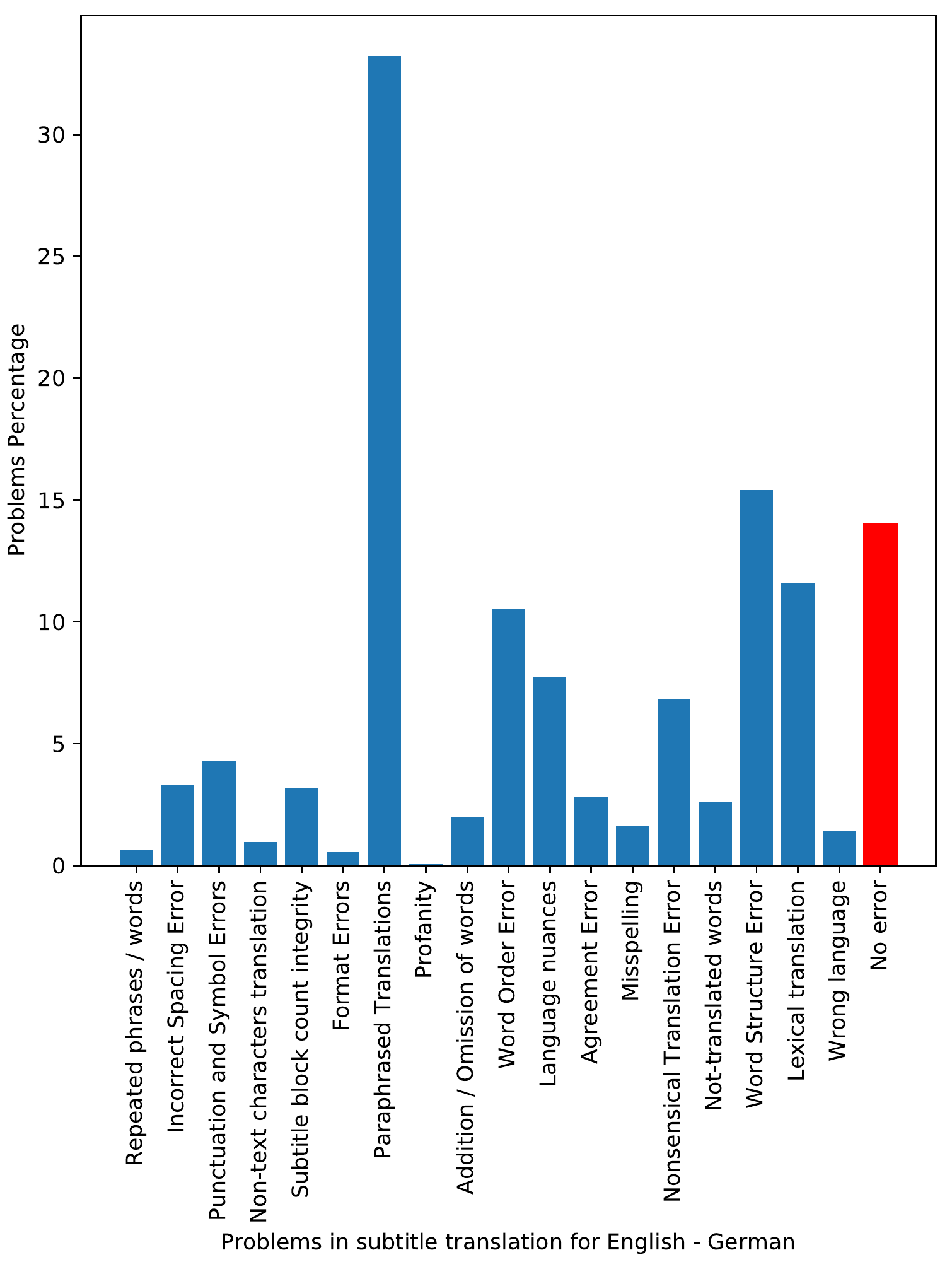}
        \caption{English to German}
        \label{subfig:en-de}
    \end{subfigure}
    ~ 
    \begin{subfigure}[b]{0.30\textwidth}
        \includegraphics[width=\textwidth]{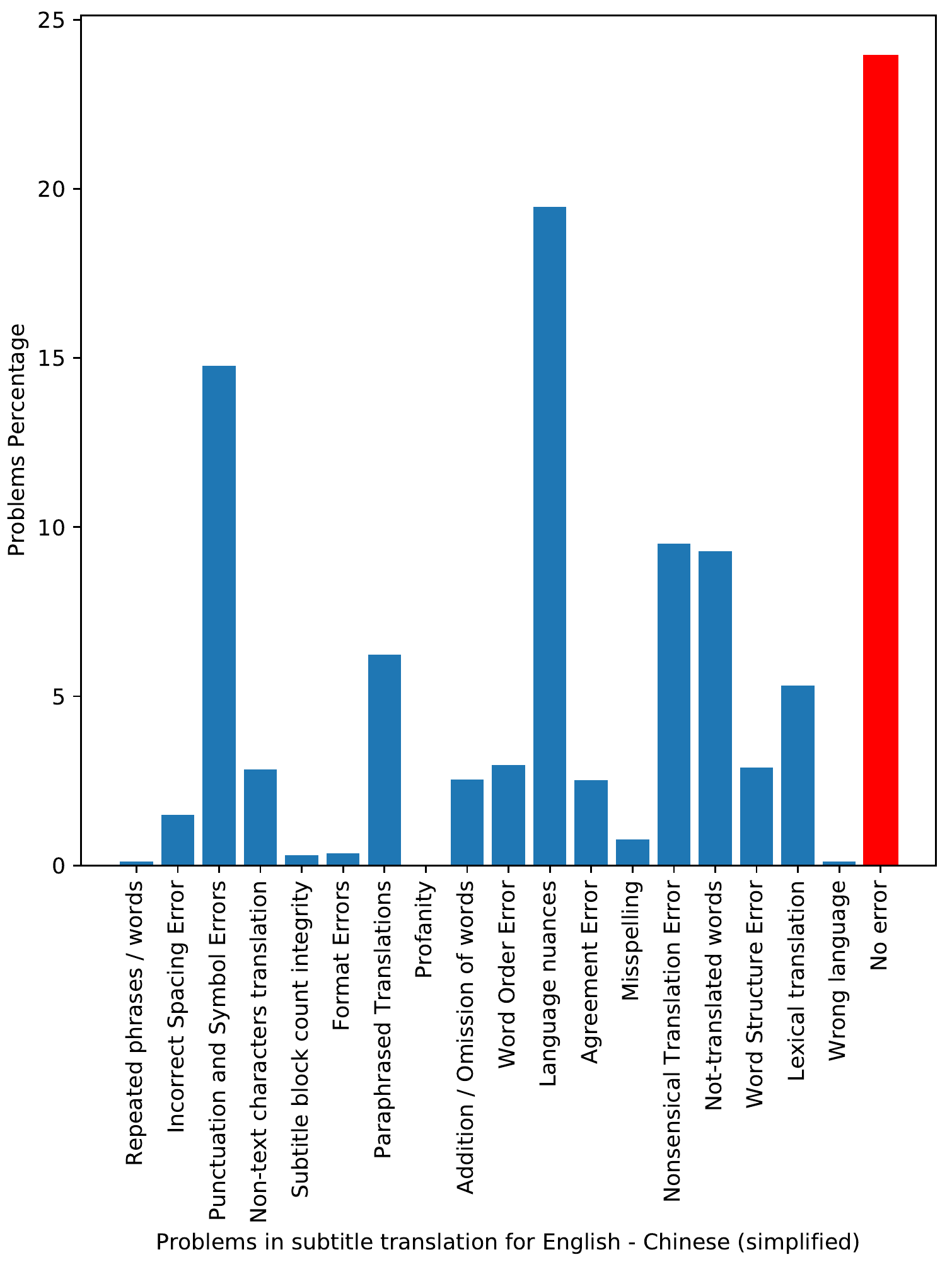}
        \caption{English to Chinese (Simplified)}
        \label{subfig: en-zh}
    \end{subfigure}
    \begin{subfigure}[b]{0.30\textwidth}
        \includegraphics[width=\textwidth]{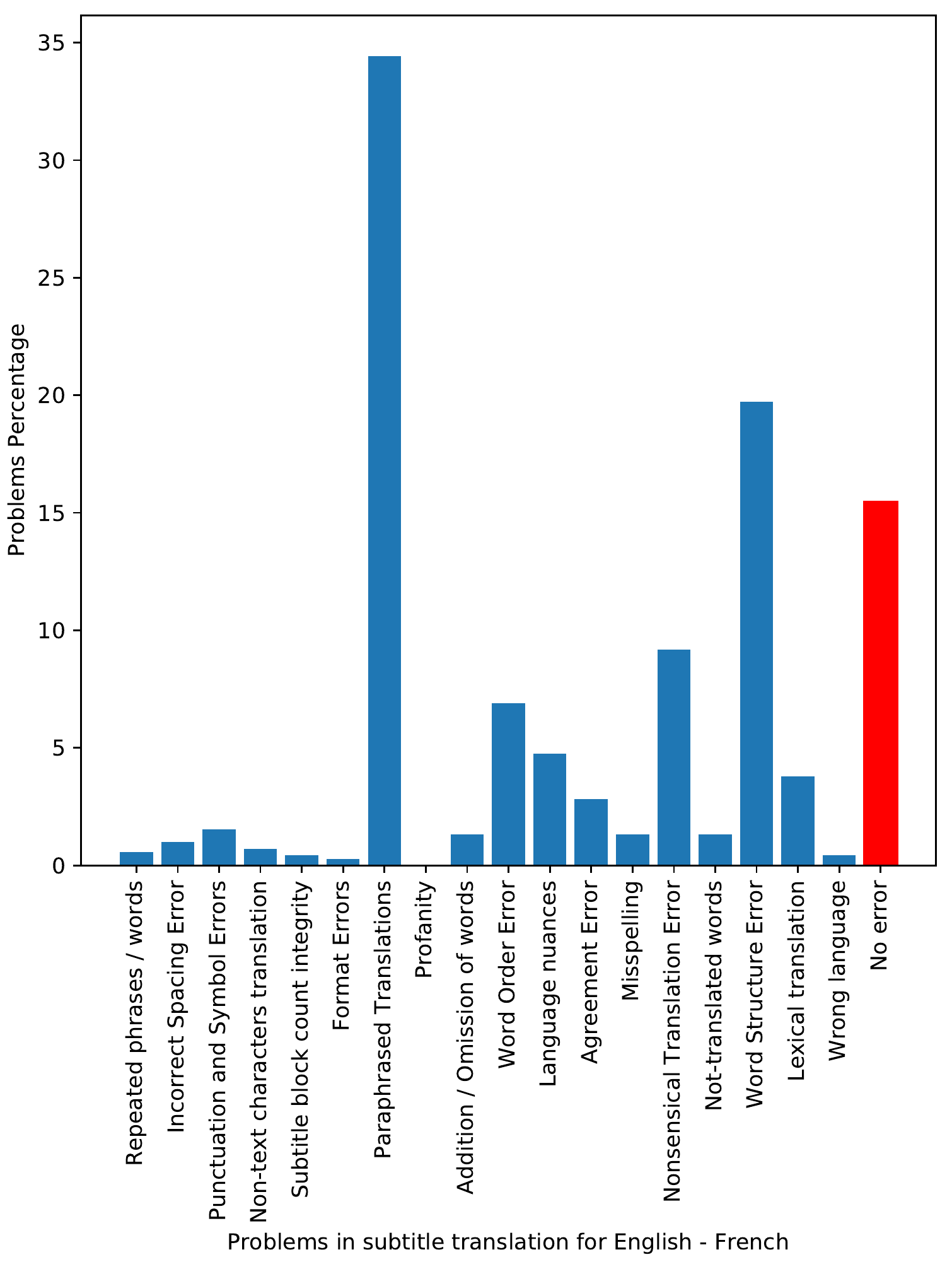}
        \caption{English to French}
        \label{subfig:en-fr}
    \end{subfigure}
    ~ 
    \begin{subfigure}[b]{0.30\textwidth}
        \includegraphics[width=\textwidth]{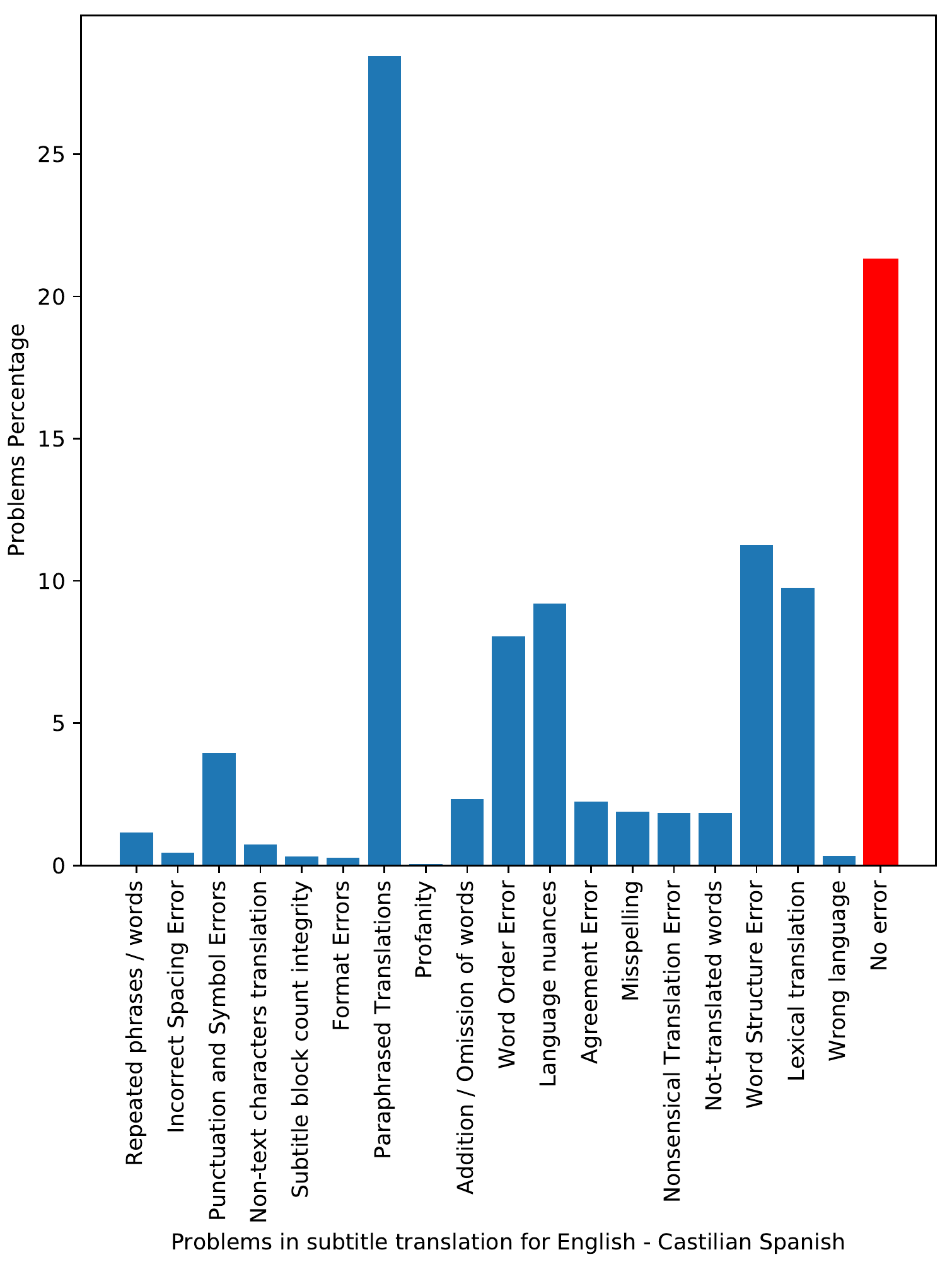}
        \caption{English to Castilian Spanish}
        \label{subfig: en-es}
    \end{subfigure}
    \begin{subfigure}[b]{0.30\textwidth}
        \includegraphics[width=\textwidth]{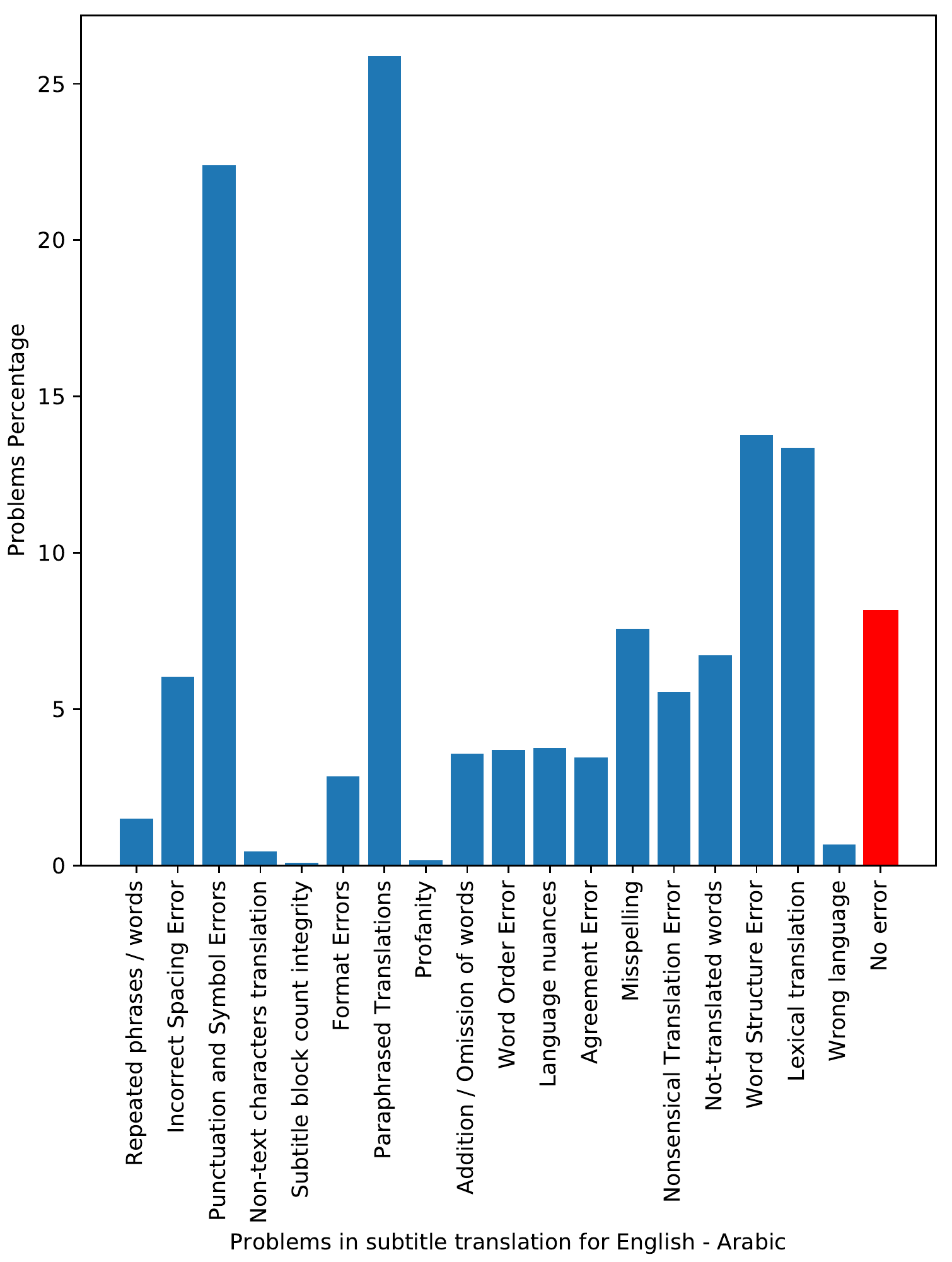}
        \caption{English to Arabic}
        \label{subfig:en-ar}
    \end{subfigure}
    ~ 
    \begin{subfigure}[b]{0.30\textwidth}
        \includegraphics[width=\textwidth]{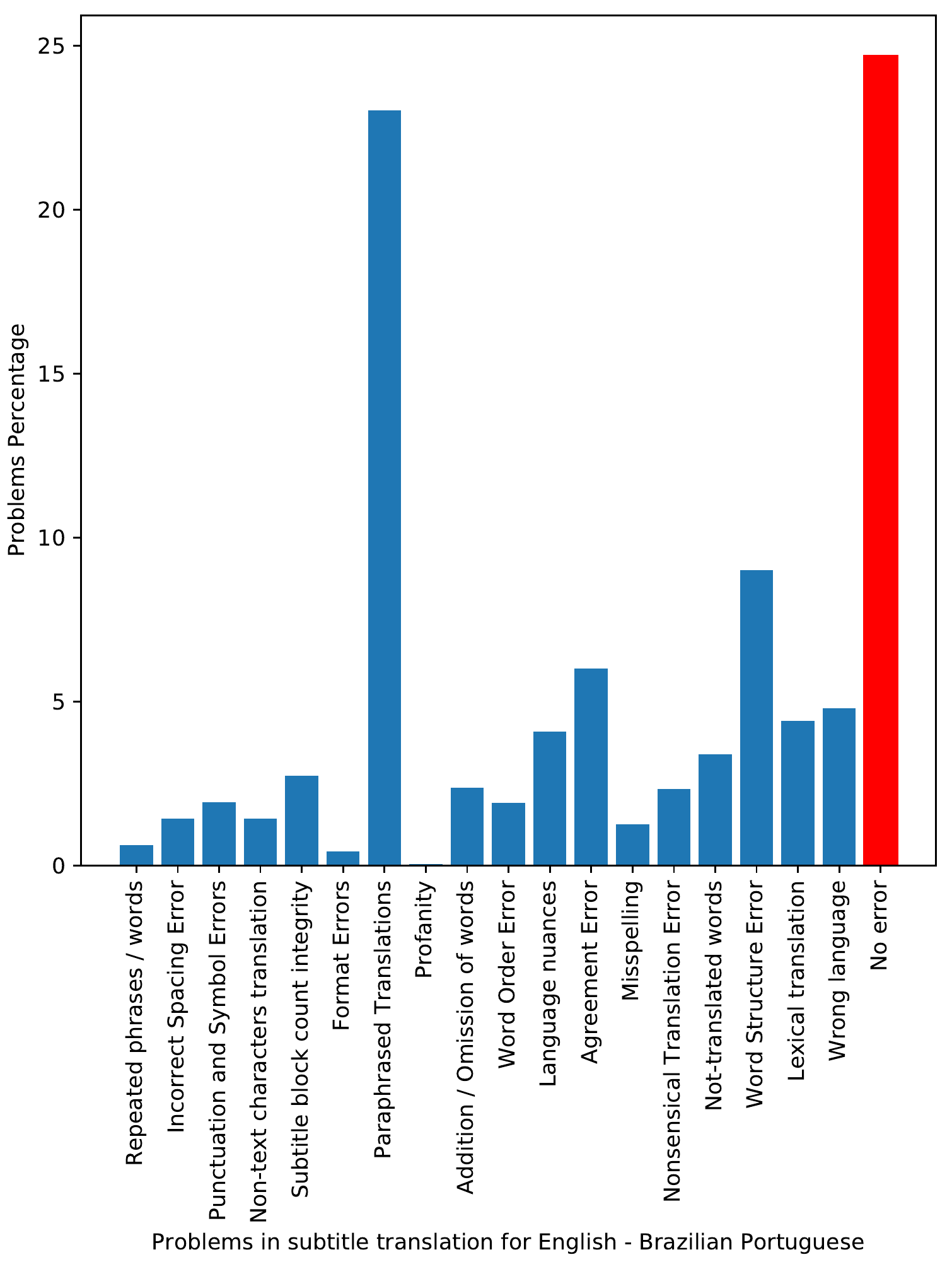}
        \caption{English to Brazilian Portuguese}
        \label{subfig: en-pt}
    \end{subfigure}
    \caption{Problem percentage for translation}
    \label{fig:enus-dede-enus-zhhans}
\end{figure*}

In this experiment we asked the professional translators to mark all the problems present in each subtitle block and provide the correct translation. 

Figure \ref{fig:enus-dede-enus-zhhans} shows the percentage of problems for all language pairs. Blue bars represent problems in translation and red bar represents percentage of blocks without these 16 errors. We observe that some of the problems are present in most of the languages. For example, Paraphrasing error is the biggest problem in all the languages except Chinese (simplified). However, certain problems are language specific and hence require specialized solutions. For example, problems of Structure Error and Word Order Error are more pronounced in German translation as compared to other languages. Non-text characters translation occurs frequently in Chinese and Arabic translations. Word Structure Error was the second biggest problem in French. For German, Spanish and Arabic, Lexical translation, was a significant problem.

\section{Conclusion}
\label{sec:Conclusion}
In this work, we explained 27 problems in automating translation for movie and TV show subtitles and share frequency of 16 key problems for six language pairs. While we do not provide possible solutions for any problems, we present an insight into the problem domain. The examples provided encourage the reader to design error-specific, language-specific and language-agnostic solutions. One can solve these problems by pre-processing the input, post-processing the MT output or by improving MT engines.
Creating one solution for all languages may not always work.





\bibliographystyle{ieeetr}
\bibliography{Bibliography-File}

\begin{thebibliography}{10}

\bibitem{Brown1990ASA}
P.~F. Brown, J.~Cocke, S.~D. Pietra, V.~J.~D. Pietra, F.~Jelinek, J.~D.
  Lafferty, R.~L. Mercer, and P.~S. Roossin, ``A statistical approach to
  machine translation,'' {\em Computational Linguistics}, vol.~16, pp.~79--85,
  1990.

\bibitem{Volk2009TheAT}
M.~Volk, ``The automatic translation of film subtitles. a machine translation
  success story?,'' {\em JLCL}, vol.~24, pp.~115--128, 2009.

\bibitem{Sennrich2010MachineTO}
R.~Sennrich, C.~Hardmeier, and F.~Tidstr{\"o}m, ``Machine translation of tv
  subtitles for large scale production,'' 2010.

\bibitem{Mller2013StatisticalMT}
M.~M{\"u}ller and M.~Volk, ``Statistical machine translation of subtitles: From
  opensubtitles to ted,'' in {\em GSCL}, 2013.

\bibitem{prabhakar2019Unsupervised}
P.~Gupta, S.~Shekhawat, and K.~Kumar, ``Unsupervised quality estimation without
  reference corpus for subtitle machine translation using word embeddings,''
  {\em IEEE 13th International Conference on Semantic Computing (ICSC)},
  pp.~32--38, jan 2019.

\bibitem{Hieber2017SockeyeAT}
F.~Hieber, T.~Domhan, M.~Denkowski, D.~Vilar, A.~Sokolov, A.~Clifton, and
  M.~Post, ``Sockeye: A toolkit for neural machine translation.,'' {\em CoRR},
  vol.~abs/1712.05690, 2017.

\bibitem{45610}
Y.~Wu, M.~Schuster, Z.~Chen, Q.~V. Le, M.~Norouzi, W.~Macherey, M.~Krikun,
  Y.~Cao, Q.~Gao, K.~Macherey, J.~Klingner, A.~Shah, M.~Johnson, X.~Liu,
  Łukasz Kaiser, S.~Gouws, Y.~Kato, T.~Kudo, H.~Kazawa, K.~Stevens, G.~Kurian,
  N.~Patil, W.~Wang, C.~Young, J.~Smith, J.~Riesa, A.~Rudnick, O.~Vinyals,
  G.~Corrado, M.~Hughes, and J.~Dean, ``Google's neural machine translation
  system: Bridging the gap between human and machine translation,'' {\em CoRR},
  vol.~abs/1609.08144, 2016.

\bibitem{Luong2015EffectiveAT}
T.~Luong, H.~Pham, and C.~D. Manning, ``Effective approaches to attention-based
  neural machine translation,'' in {\em EMNLP}, 2015.

\bibitem{Britz2017MassiveEO}
D.~Britz, A.~Goldie, M.-T. Luong, and Q.~V. Le, ``Massive exploration of neural
  machine translation architectures,'' {\em CoRR}, vol.~abs/1703.03906, 2017.

\bibitem{Vaswani2017AttentionIA}
A.~Vaswani, N.~Shazeer, N.~Parmar, J.~Uszkoreit, L.~Jones, A.~N. Gomez,
  L.~Kaiser, and I.~Polosukhin, ``Attention is all you need,'' in {\em NIPS},
  2017.

\bibitem{le2017improving}
A.~N. Le, A.~Martinez, A.~Yoshimoto, and Y.~Matsumoto, ``Improving sequence to
  sequence neural machine translation by utilizing syntactic dependency
  information,'' in {\em Proceedings of the Eighth International Joint
  Conference on Natural Language Processing (Volume 1: Long Papers)}, vol.~1,
  pp.~21--29, 2017.

\bibitem{Marco2017SimpleCS}
M.~W.-D. Marco, ``Simple compound splitting for german,'' in {\em MWE@EACL},
  2017.

\bibitem{Popovic2006StatisticalMT}
M.~Popovic, D.~Stein, and H.~Ney, ``Statistical machine translation of german
  compound words,'' in {\em FinTAL}, 2006.

\bibitem{Escartn2014GermanCA}
C.~P. Escart{\'i}n, S.~Peitz, and H.~Ney, ``German compounds and statistical
  machine translation. can they get along?,'' in {\em MWE@EACL}, 2014.

\bibitem{lakew2018neural}
S.~M. Lakew, A.~Erofeeva, and M.~Federico, ``Neural machine translation into
  language varieties,'' {\em arXiv preprint arXiv:1811.01064}, 2018.

\bibitem{romero2009more}
P.~Romero-Fresco, ``More haste less speed: Edited versus verbatim respoken
  subtitles.,'' {\em Vigo International Journal of Applied Linguistics},
  vol.~6, 2009.

\bibitem{Anastasiou2010IdiomTE}
D.~Anastasiou, ``Idiom treatment experiments in machine translation,'' 2010.

\bibitem{knowles2018context}
R.~Knowles and P.~Koehn, ``Context and copying in neural machine translation,''
  in {\em Proceedings of the 2018 Conference on Empirical Methods in Natural
  Language Processing}, pp.~3034--3041, 2018.

\bibitem{vilaCabrera2015AnAO}
J.~J. {\'A}vila-Cabrera, ``An account of the subtitling of offensive and taboo
  language in tarantino ’ s screenplays,'' 2015.

\bibitem{Galinskaya2014MeasuringTI}
I.~Galinskaya, V.~Gusev, E.~Mescheryakova, and M.~Shmatova, ``Measuring the
  impact of spelling errors on the quality of machine translation,'' in {\em
  LREC}, 2014.

\bibitem{Luong2015AddressingTR}
T.~Luong, I.~Sutskever, Q.~V. Le, O.~Vinyals, and W.~Zaremba, ``Addressing the
  rare word problem in neural machine translation,'' in {\em ACL}, 2015.

\bibitem{Tu2016ModelingCF}
Z.~Tu, Z.~Lu, Y.~Liu, X.~Liu, and H.~Li, ``Modeling coverage for neural machine
  translation,'' in {\em ACL}, 2016.

\bibitem{scripture1922some}
M.~K. Scripture, ``Some theories concerning stuttering and stammering,'' {\em
  Quarterly Journal of Speech}, vol.~8, no.~2, pp.~145--155, 1922.

\bibitem{Wees2018EvaluationOM}
M.~van~der Wees, A.~Bisazza, and C.~Monz, ``Evaluation of machine translation
  performance across multiple genres and languages,'' in {\em LREC}, 2018.

\end{thebibliography}

\end{document}